\documentclass[11pt,letterpaper]{article}
\usepackage{float}
\newcommand {\be}{\begin{equation}}
\newcommand {\ee}{\end{equation}}
\newcommand {\bea}{\begin{eqnarray}}
\newcommand {\eea}{\end{eqnarray}}

\usepackage[dvips]{graphicx}
\graphicspath{{../eps/}}
\DeclareGraphicsExtensions{.eps}
\usepackage{epsfig,color}
\usepackage{booktabs}
\usepackage{epstopdf}
\usepackage{amssymb}
\usepackage{mathtools,cite}
\usepackage{xspace} 
\usepackage{algorithmic}

\usepackage{array}

\usepackage{mdwmath}
\usepackage{mdwtab}
\usepackage{bm}
\usepackage{mathrsfs}
\usepackage{epsfig}
\usepackage{subfigure}
\usepackage{algorithmic}
\usepackage{amsmath}
\usepackage{mathrsfs}
\usepackage{cleveref}  
\usepackage[section]{placeins}
\begin{document}

%\draft
\begin{centering}
%{\large \bf
%Phases of learning dynamics in stochastic gradient descent with and without mislabeled data}\\
%{\large \bf
%Dynamic phases of learning with and without mislabeled data}\\
{\large \bf
Phases of learning dynamics in artificial neural networks: with or without mislabeled data}\\
\vspace{0.5cm}
Yu Feng and Yuhai Tu\\
\vspace{0.25cm}
IBM T. J. Watson Research Center\\
Yorktown Heights, NY10598\\
%(\today{})

%\email{yuhai@us.ibm.com}

\end{centering}
\vspace{0.5cm}

\begin{center}
{\bf Abstract}
\end{center}
Despite tremendous success of deep neural network in machine learning, the underlying reason for its superior learning capability remains unclear. Here, we present a framework based on statistical physics to study dynamics of stochastic gradient descent (SGD) that drives learning in neural networks. By using the minibatch gradient ensemble, we construct order parameters to characterize dynamics of weight updates in SGD. In the case without mislabeled data, we find that the SGD learning dynamics transitions from a fast learning phase to a slow exploration phase, which is associated with large changes in order parameters that characterize the alignment of SGD gradients and their mean amplitude. In the more complex case with randomly mislabeled samples, SGD learning dynamics falls into four distinct phases. The system first finds solutions for the correctly labeled samples in phase I, it then wanders around these solutions in phase II until it finds a direction to learn the mislabeled samples during phase III, after which it finds solutions that satisfy all training samples during phase IV. Correspondingly, the test error decreases during phase I and remains low during phase II; however, it increases during phase III and reaches a high plateau during phase IV. The transitions between different phases can be understood by changes of order parameters that characterize the alignment of the mean gradients for the two datasets (correctly and incorrectly labeled samples) and their (relative) strength during learning. We find that individual sample losses for the two datasets are most separated during phase II, which leads to a cleaning process to eliminate mislabeled samples for improving generalization. Overall, we believe that the approach based on statistical physics and stochastic dynamical systems theory provides a promising framework to describe and understand learning dynamics in neural networks, which may also lead to more efficient learning algorithms.    

\newpage

\section{Introduction: Learning as a stochastic dynamical system}

Modern artificial neural network-based algorithms, in particular deep learning neural network (DLNN) ~\cite{LeCun2015Deep,goodfellow2016deep}, have enjoyed a long string of tremendous successes in achieving human level performance in image recognition~\cite{he2016deep}, machine translation~\cite{wu2016google}, games~\cite{alpha-go}, and even solving longstanding grand challenge scientific problems such as protein folding~\cite{alpha-fold}. However, despite DLNN's successes, the underlying mechanism of how they work remains unclear. For example, one key ingredient for the powerful DLNN is a relatively simple iterative method called stochastic gradient descent (SGD)~\cite{Robbins_1951,SGD2010Bottou}. However, the reason why SGD is so effective in finding highly generalizable solutions in a high dimensional nonconvex loss function landscape remains unclear. The random elements due to subsampling in SGD seems key for learning, yet the inherent noise in SGD also makes it difficult to understand. 

From thermodynamics and statistical physics, we know that physical systems with many degrees of freedom are subject to stochastic fluctuations, e.g., thermal noise that drives Brownian motion, and powerful tools have been developed for understanding collective behaviors in stochastic processes~\cite{Kampen}. In this paper, we propose to consider the SGD based learning process as a stochastic dynamical system and to investigate the SGD-based learning dynamics by using concepts and methods from statistical physics. 

In an artificial neural network (ANN), the model is parameterized by its weights represented as a $N_p-$dimensional vector: $w=(w_1,w_2,.....,w_{N_p})$ where $N_p$ is the number of parameters (weights). For supervised learning, there is a set of $N$ training samples each with an input vector $X_k$ and a correct output vector $Z_k$ for $k=1,2,...,N$. For each input $X_k$, the learning system predicts an output vector $Y_k=G(X_k,w)$, where the output function $G$ depends on the architecture of the NN as well as its weights $w$. The goal of learning is to find the weight parameters to minimize the difference between the predicted and correct output characterized by an overall loss function (or energy function):
\begin{equation}
    L(w)=N^{-1}\sum_{k=1}^{N} l_k ,
\end{equation}
where $l_k = d(Y_k,Z_k)$ is the loss for sample $k$ that measures of distance between $Y_k$ and $Z_k$. A popular choice for $d$ is the cross-entropy loss, which is what we use in this paper. 

One learning strategy is to update the weights by following the gradient of $L$ directly. However, this direct gradient descent (GD) scheme is computationally prohibitive for large datasets and it also has the obvious shortfall of being trapped by local minima or saddle points. SGD was first introduced to circumvent the large dataset problem by updating the weights according to a subset (minibatch) of samples randomly chosen at each iteration~\cite{Robbins_1951}. Specifically, the change of weight $w_i$ $(i=1,2,...,N_p)$ for iteration $t$ in SGD is given by:
\begin{equation}
\Delta w_i (t) = -\alpha \frac{\partial L^{\mu(t)}(w)}{\partial w_i},
\label{backprop}
\end{equation}
where $\alpha$ is the learning rate and $\mu(t)$ represents the random minibatch used for iteration $t$. The mini loss function (MLF) for minibatch $\mu$ of size $B$ is defined as:
\begin{equation}
    L^\mu(w) =B^{-1} \sum_{l=1}^{B} d(Y_{\mu_l},Z_{\mu_l}),
\end{equation} 
where $\mu_l$ ($l=1,2,..,B$) labels the $B$ randomly chosen training samples. 

Besides the computational advantage of SGD, the inherent noise due to random subsampling in SGD allows the system to escape local traps. Noise in SGD comes from the difference of the minibatch loss function $L^\mu$ and the whole batch loss function $L$: $\delta L^{\mu} \equiv L^{\mu}-L$. By taking the continuous time approximation in Eq. (\ref{backprop}), the SGD learning dynamics can be described by a Langevin equation:
\begin{equation}
\frac{d w}{dt}= -\alpha \nabla_{w} L+\eta,
\label{Langevin}
\end{equation}
where the first term on the right hand side (RHS) of Eq.~\ref{Langevin} is the usual deterministic gradient descent term, and the second term  corresponds to the SGD noise defined as: $\eta \equiv -\alpha \nabla \delta L^\mu$. The SGD noise has zero mean $\langle \eta \rangle_\mu = 0$ and its strength is characterized by the noise matrix:
$
\Delta_{ij}\equiv \langle \eta_i \eta_{j} \rangle = \alpha^2 C_{ij}$,   
where the co-variance matrix ${\bf C}$ can be written as:
\begin{equation}
 C_{ij}\equiv \langle \frac{\partial \delta L^{\mu}}{\partial w_i} \frac{\partial \delta L^{\mu}}{\partial w_j} \rangle _\mu = \langle \frac{\partial L^{\mu}}{\partial w_i} \frac{\partial  L^{\mu}}{\partial w_j} \rangle _\mu - \frac{\partial L}{\partial w_i}\cdot \frac{\partial L}{\partial w_j}.
 \label{CL}
 \end{equation}
 
  %The SGD Langevin equation was was first studied by using the Langevin equation approach by Chaudhari and Soatto~\cite{Chaudhari_2018}.
According to Eq.~\ref{Langevin}, the SGD based learning dynamics can be considered as stochastic motion of a ``learning particle" ($w$) in the high-dimensional weight space. In physical systems that are in thermal equilibrium, their stochastic dynamics can also be described by Langevin equations with the same deterministic term as in Eq.~\ref{Langevin} but with a much simpler noise term that describes the isotropic and homogeneous thermal fluctuations. Indeed, as first pointed out by Chaudhari and Soatto~\cite{Chaudhari_2018}, the SGD noise is neither isotropic nor homogeneous in weight space. In this sense, the SGD noise is highly nonequilibrium. As a result of the nonequilibrium SGD noise, the steady state distribution of weights is not the Boltzmann distribution as in equilibrium systems, and SGD dynamics exhibits much richer behaviors than simply minimizing a global loss function (free energy).   

How can we understand SGD-based learning in ANN? Here, we propose to bring useful concepts and tools from statistical physics~\cite{forster2018hydrodynamic} and stochastic processes~\cite{Kampen} to bear on characterizing and investigating the SGD learning process/dynamics. In the rest of this paper, we describe a systematic way to characterize SGD dynamics based 
on order parameters that are defined over the minibatch gradient ensemble. We show how this approach allows us to identify and understand various phases in the learning process without and with labeling noise, which may lead to useful algorithms to improve generalization in the presence of mislabeled data. Throughout our study, we use realistic but simple datasets to demonstrate the principles of our approach with less attention paid to the absolute performance.

\section{Characterizing SGD learning dynamics: the minibatch gradient ensemble and order parameters}

%For studying ``dynamics " of the artificial learning process, we can label the sequence of the mini-batch presented to the learner by the `` learning time" $t$. For off-line learning, the sequence $\mu(t)$ may be random like in SGD or with some ordering like in curriculum learning where the samples are presented in an order according to the ``complexity" of the sample/task. For on-line learning, $t$ actually has a physical meaning as it represents the order samples are presented in the on-line setting. The learner at time $t$ has access to the current sample $\mu(t)$ and may be some "old" samples stored in a finite buffer that can be used to help training with the new sample like in various online algorithms with replay. 

%The stochastic nature of the learning dynamics comes from the random sub-sampling (minibatch) from the whole training dataset (whole batch).
To characterize the stochastic learning dynamics in SGD, we introduce the concept of minibatch ensemble $\{\mu\}$ where each member of the ensemble is a minibatch with $B$ samples chosen randomly from the whole training dataset (size $N$). Based on the minibatch ensemble, we can define an ensemble of minibatch loss functions $L^\mu$ or equivalently an ensemble of gradients $\{g^\mu (\equiv -\nabla L^{\mu}(w))\}$ at each weight vector $w$. %At a given time $t$, a member of the minibatch ensemble ($\mu(t)$) is chosen randomly, and the weight dynamics in SGD follows the gradient $g^{\mu(t)}$.

The SGD learning dynamics is fully characterized by statistical properties of the gradient ensemble in weight space $\{g^\mu (w)\}$. At each point in weight space, the ensemble average of the minibatch gradients is the gradient over the whole dataset: $g(w) \equiv \langle g^\mu (w) \rangle_\mu (=\nabla L(w))$, and fluctuations of the gradients around their mean give rise to the noise matrix (Eq.~\ref{CL}). To measure the alignment among the minibatch gradients, we define an alignment parameter $R$:
\begin{equation}
    R(w)\equiv \langle \hat{g}^\mu(w) \cdot \hat{g}^\nu (w) \rangle _{\mu,\nu},
\end{equation}
where $\hat{g}^\mu =g^\mu/\|g^\mu\|$ is the unit vector in gradient direction $g^\mu$. The alignment parameter is the cosine of the relative angle between two gradients averaged over all pairs of minibatches $(\mu,\nu)$ in the ensemble.

To analyze the gradient fluctuations in different directions, we can project the minibatch gradient $g^\mu$ onto the mean $g$ and write it as:
\begin{equation}
    g^\mu = g^\mu_{\bot} +\lambda_\mu g,
\end{equation}
where $\lambda_\mu=(g^\mu \cdot g)/\|g\|^2$ is the projection constant and $g^\mu_{\bot}$ is the residue gradient perpendicular to $g$: $g^\mu_{\bot} \cdot g =0$. In analogy to kinetic energy, we use the square of the gradient to measure the learning activity. The ensemble averaged activity $(A)$ can be split into two parts:
\begin{equation}
    A\equiv \langle \|g^\mu\|^2\rangle_\mu =\langle \|g^\mu_{\bot}\|^2\rangle_\mu +\langle \lambda^2_\mu\rangle_\mu \|g\|^2 \equiv A_{\bot}+A_{\|},
\end{equation}
where $A_{\|}$ and $A_{\bot}$ represent activities along the mean gradient and orthogonal to it, respectively. 

The total variance $D$ of fluctuations in all directions is the trace of the co-variance matrix ${\bf C}$:
\begin{equation}
    D\equiv Tr({\bf C})=\sum_i C_{ii} =A_{\bot} + D_{\|},
\end{equation}
where $D_{\|}=\sigma_{\lambda}^2\|g\|^2$ is the variance along the direction of the batch gradient $g$ with $\sigma^2_\lambda\equiv \langle \lambda^2_\mu\rangle_\mu -1$ the variance of $\lambda_\mu$ (Note that $\langle \lambda_\mu\rangle_\mu =1$ by definition); $A_{\bot}$ is the total variance in the orthogonal directions. The mean learning activity can be written as: $A=A_0 +A_{\bot} + D_{\|}$, where $A_0\equiv \|g\|^2$ represents the directed activity along the mean gradient direction; $A_{\bot}$ and $D_{\|}$ represent the diffusive search activities along the directions orthogonal and parallel to the mean gradient, respectively.

All these quantities ($A$, $A_0$, $R$, $\sigma^2_\lambda$) depend on the weights ($w$). Along a SGD learning trajectory in weight space, we can evaluate these order parameters and their relative values at any given time $t$ to characterize different phases of the SGD learning dynamics. For example, we use $A$ and $A_0$ to measure the total learning activity and the activity along the mean gradient direction respectively. The alignment among different minibatch gradients is measurement by $R$, which is related to the fractional aligned activity $ A_0/A$. The fluctuations of the minibatch gradients projected onto the mean gradient is measured by $\sigma^2_\lambda$. In our previous work~\cite{feng2020neural}, we used time averaging to approximate some of these order parameters for computational convenience. However, properties of the SGD dynamics at any given point in weight space are precisely defined by these ensemble averaged order parameters, which is used hereafter. 

As mentioned before, the SGD noise is anisotropic and varies in weight space. The positive-definite eigenvalue  $e_l$ of the symmetric co-variance matrix ${\bf C}$ is the noise strength in the corresponding eigen-direction ($l=1,2,...,N_p$ with $N_p$ the number of weights or the dimension of the weight space). The overall noise strength $D=Tr({\bf C})=\sum_{l=1}^{N_p} e_l$ describes the total search activity, and the eigenvalue spectrum $\{e_l, \; l =1,2,...,N_p\}$ tells us how much of the total search activity is spent in each eigen-direction. From the noise spectrum, we can define an effective dimension of search activity $D_s(w)$ as the number of dimensions wherein the variance in the subspace of parameters account for certain large percentage (e.g., $90\%$) of the total variance $D$. 

\section{Phases of SGD learning dynamics without mislabeled data}

\begin{figure}[htbp]
\centering
\includegraphics[width=0.75\linewidth]{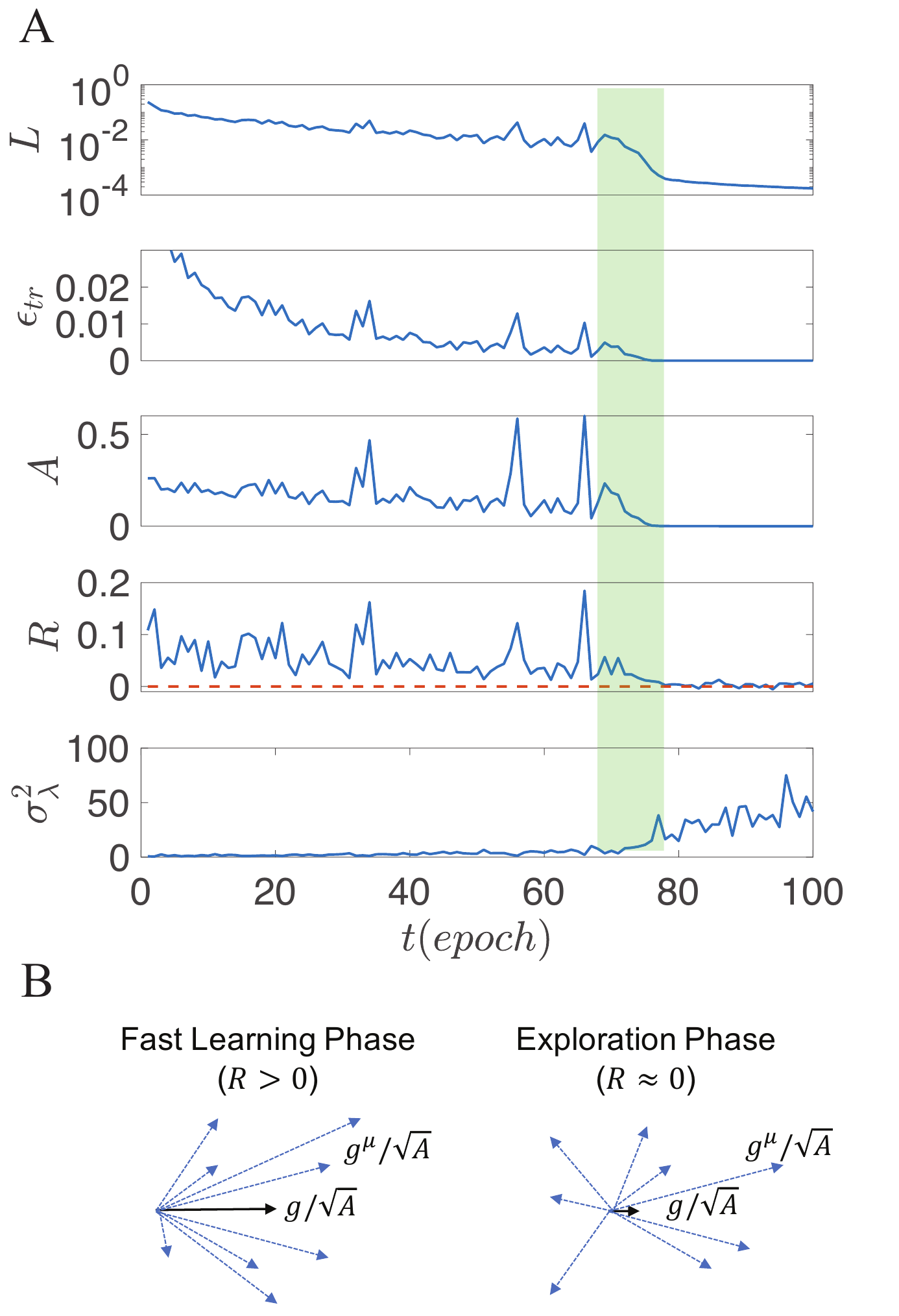}
\caption{Two phases of learning without labeling noise. (A) Training loss $L$, training error $\epsilon_{tr}$, and order parameters $A$, $R$, and $\sigma^2_\lambda$ versus (training) time. The fast learning phase corresponds to a directed (finite $R>0$, $\sigma^2_\lambda\sim 1$) and fast (large $A$) motion in weight space; the exploration phase corresponds to a diffusive ($R\approx 0$, $\sigma^2_\lambda\gg 1$) and slow (small $A$) motion in weight space. The dotted line shows $R=0$. The green bar highlights the transition region. MNIST data and an all connected network with 2 hidden layers ($30\times 30$) are used here. (B) Illustration of the normalized minibatch gradient ensemble (blue dotted arrows) and their means (black solid arrows) in the two learning phases.} %{\color{red} Depending on your simulation results, this figure needs to be changed by using the ensemble average .... Also, I think we should add an additional panel to look at the distribution of individual losses at different times like in the noise-label case.}  } 
\label{pt} 
\end{figure}

We first study the learning dynamics without mislabeled data, e.g., the original MNIST dataset. As shown in Fig.~\ref{pt}, dynamics of the overall loss function $L$ suggests that there are two phases in learning. There is an initial fast learning phase where $L$ decreases quickly followed by an exploration phase where the training error $\epsilon_{tr}$ reaches $0$ (or nearly $0$) while $L$ still decreases but much more slowly. %Due to its slow dynamics, the exploration phase can be considered as in a quasi-steady-state.  
These two learning phases exist independent of hyperparameters (e.g., $\alpha$ and $B$) and network architectures (all connected network or $CNN$) used for different datasets (e.g., MNIST and CIFAR). %  as shown in Fig.~\ref{pt}, where the transition region between the two phases are highlighted. 
The weights reached in the exploration phase can be considered as solutions of the problem given that the training error vanishes. % and the test error seems to reach a low steady state value in the exploration phase.

Dynamics of the order parameters $A(t)$, $R(t)$, and $\sigma^2_\lambda$ along the trajectory can be used to characterize and understand the two phases. As shown Fig.~\ref{pt}(A), in the beginning of the learning process, the learning activity $A$ is relatively large and the alignment parameter $R$ is finite. In this initial phase of learning, the minibatch gradients have a high degree of alignment resulting to a strongly directed motion of the weight particle and a fast decrease of $L$ towards a solution region in the weight space with low $L$ and zero training error $\epsilon_{tr}$. In the exploration phase, the average learning activity $A$ becomes much smaller while the average alignment parameter $R$ becomes close to zero. This means that the motion of the weight particle becomes mostly diffusive (weakly directed) and the decrease of $L$ slows. This diffusive motion of weights allows the system to explore the solution space. The transition from a directed motion to a diffusive motion is also reflected in the large increase of the variance $\sigma^2_\lambda$ at the transition. Due to the finite size of the system, the transition is not infinitely sharp as phase transition in physical systems in thermodynamic limit (infinite system limit). As shown in Fig.~\ref{pt}(A), the training error $\epsilon_{tr}$ becomes zero during the transition regime and it stays zero in the exploration phase. These results confirm our previous study that used the time-averaged ordered parameters~\cite{feng2020neural}. Key differences between the two phases in terms of alignment of minibatch gradients and mean gradient strength are illustrated in Fig.~\ref{pt}(B).   

\begin{figure}[htbp]
\centering
\includegraphics[width=0.6\linewidth]{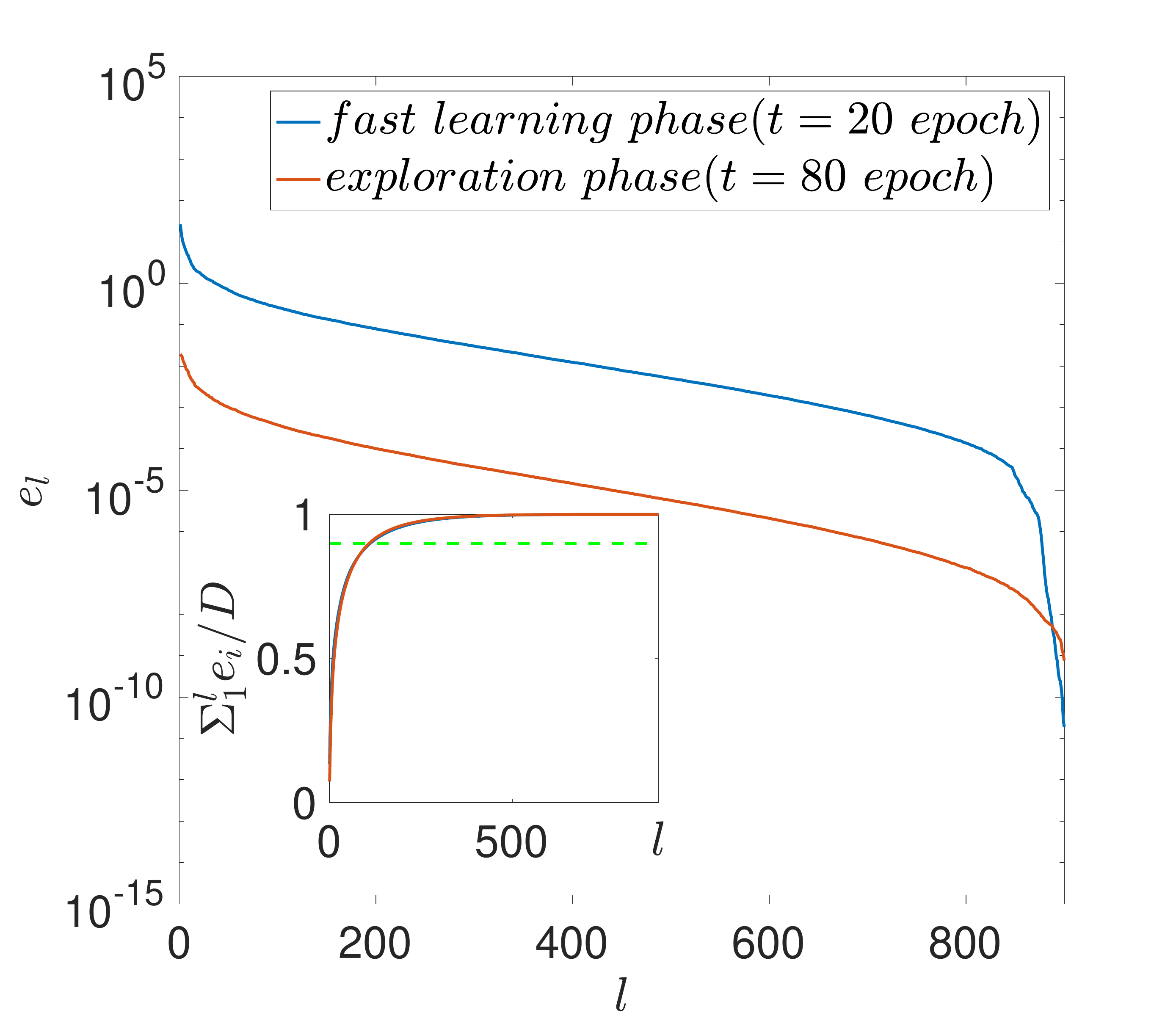}
\caption{The noise spectra, i.e., rank ordered eigenvalues $\{e_l,\; l=1,2..,N_p\}$ in the fast learning phase (black) and the exploration phase (red) . The inset shows the normalized accumulated variance $D^{-1}\sum_{i=1}^{l} e_i$. The two spectra are similar except for their total variance $D$. The effective dimension $D_s\sim 110$, which is much smaller than the number of parameters ($N_p=900$), is roughly the same in both phases. Data and network used here are the same as in Fig.~\ref{pt}.} 
\label{noise_spec} 
\end{figure}

We have also studied the noise spectra in the two phases. As shown in Fig.~\ref{noise_spec}, unlike isotropic thermal noise, the SGD noise has a highly anisotropic structure with most of its variance (strength) concentrated in a relatively small number of directions. The normalized noise spectra are similar in both phases and the total noise strength (variance) $D$ is much higher in the fast learning phase. The effective dimension defined as the number of directions that contains $90 \%$ of the total variance is $D_s\sim 110$, which is much smaller than the number of weighs (parameters) and remains roughly constant as the number of parameters increases.    %The most prominent difference between noise in these two phases seems to be their total strength $D$.  

\section{Phases of SGD learning dynamics in the presence of mislabeled data}

There has been much interest in deep learning in the presence of mislabeled data. This is triggered by a recent study~\cite{Zhang_2018} in which the authors showed that random labels can be easily fitted by deep networks in the over-parameterized regime and such overfitting destroys generalization. Here, we report some new results by using the dynamical systems approach developed in previous sections to study SGD learning dynamics with labeling noise. 

In a dataset with $N_c$ correctly labeled training samples and $N_w$ incorrectly (randomly) labeled samples, the overall loss function $L$ consists of two parts, $L_c$ and $L_w$, from the correctly-labeled samples and the randomly labeled samples, respectively:
\begin{equation}
L=(1-\rho)L_c +\rho L_w=N^{-1}\Big[\sum_{k=1}^{N_c} l_k +\sum_{k=1}^{N_w}\tilde{l}_{k}\Big],
\end{equation}
where $N=N_c+N_w$ is the total number of training samples and $\rho=N_w/N$ is the fraction of mislabeled samples. The loss function for a correctly labeled sample is the cross entropy $l$ between the output $Y_k(X_k,w)$ of the network with weight vector $w$ and the correct label vector $Z_k$: $l_k=l(Y_k,Z_k)$; whereas the loss function for a mislabeled sample is: $\tilde{l}_k =l(Y_k,Z^r_k)$ where $Z^r_k$ is a random label vector.

\begin{figure}[ht]
\centering
\includegraphics[width = 1.0\linewidth]{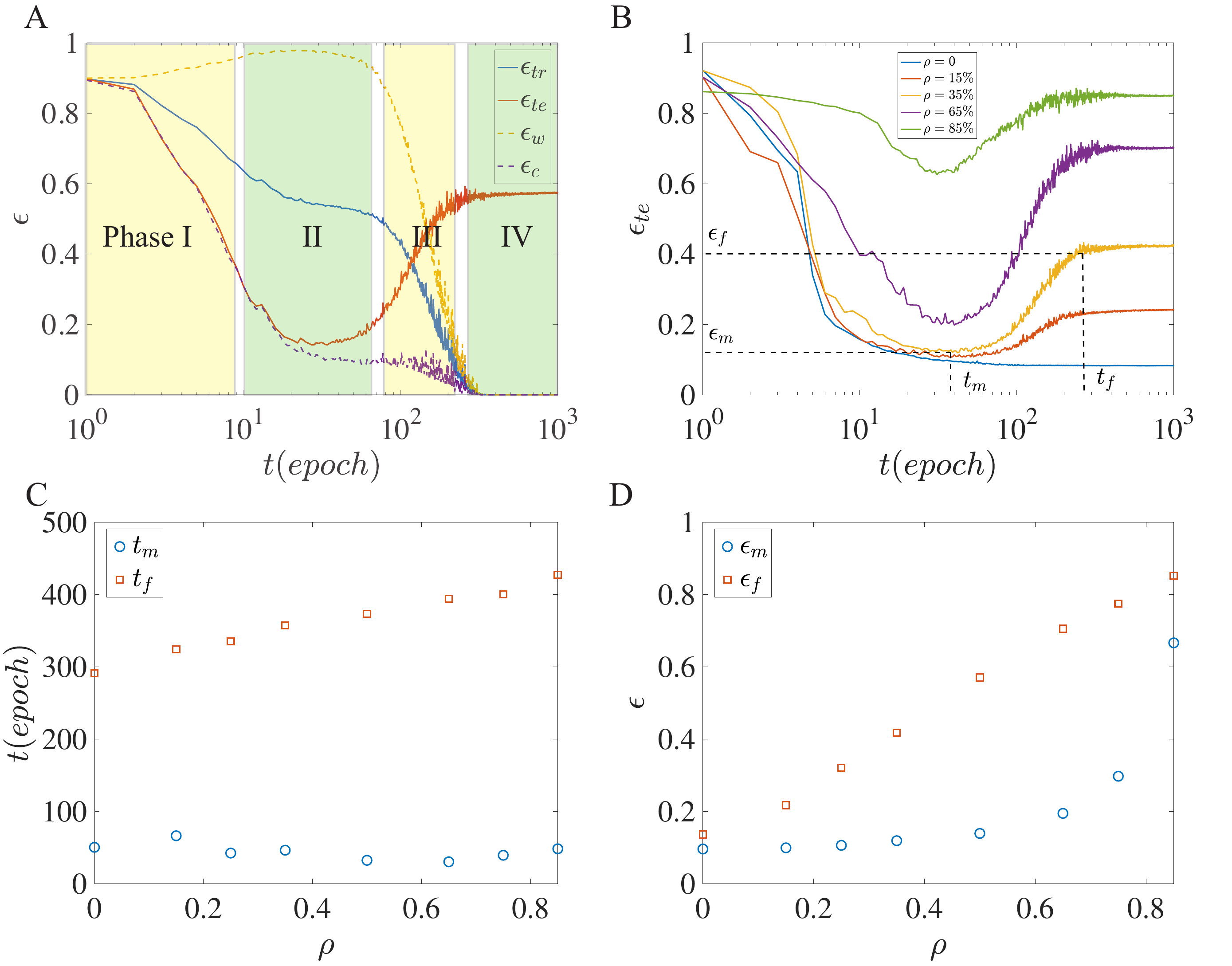} 

\caption{Learning dynamics in the presence of labeling noise. (A) The training error $\epsilon_{tr}$, the test error $\epsilon_{te}$, the training error for correctly labeled data $\epsilon_{c}$, the training error for mislabeled data $\epsilon_{w}$ are shown for a subset of MNIST data with 400 samples per digit and a fully connected network with two hidden layer (50 hidden units per layer). SGD hyper-parameters: $B=25$, $\alpha =0.01$. (B) $\epsilon_{te}$ dynamics for different values of $\rho$. (C) The dependence of the time scales ($t_m$ and $t_f$) on $\rho$. (D) The dependence of the minimum and final test errors ($\epsilon_m$ and $\epsilon_f$) on $\rho$.}
\label{rho}
\end{figure}

We did experiment on the MNIST and CIFAR10 with different fractions of mislabeled data ($\rho$). As shown in Fig.~\ref{rho}(A) for MNIST, the whole learning process can be divided into 4 phases (study of the CIFAR10 dataset shows similar results):
\begin{itemize}
\item Phase I: During this initial fast learning phase ($0-10$ epoch in Fig.~\ref{rho}(A)), the test error $\epsilon_{te}$ decreases quickly as the system learns the correctly labeled data. The error $\epsilon_c$ from the correctly labeled training data follows the exact same trend as $\epsilon_{te}$ and the error $\epsilon_w$ from the mislabeled training data actually increases slightly, which indicates that learning in phase I is dominated by the correctly labeled training data.  
\item Phase II: After the initial fast learning phase, the test error $\epsilon_{te}$ stays roughly the same during phase II ($10-70$ epoch in Fig.~\ref{rho}(A)). Both $\epsilon_w$ and $\epsilon_c$ remains flat, which indicates that learning activities for the correct and incorrect samples are balanced during phase II. This can also be seen in the plateau in the total training error $\epsilon_{tr}=\epsilon_c +\epsilon_w$.   
\item Phase III: At the end of phase II ($\sim 70$ epoch), the test error $\epsilon_{te}$ starts to increase quickly while the training errors for both the correct and the incorrect training data ($\epsilon_c$, $\epsilon_w$) decreases to zero during phase III ($70-200$ epoch). During phase III, the system finally manages to find (learn) a solution that satisfies both the correct and incorrect training data.
\item Phase IV: Phase IV corresponds to the slow exploration phase after the system reaches the solution space for the whole dataset. The test error reaches a high plateau in phase IV.  
\end{itemize}

The four distinct phases in the presence of labeling noise and the corresponding ``U"-shaped behavior in test error are general for a wide range of noise level ($\rho$), see Fig.~\ref{rho}(B). Quantitatively, dynamics of the test error $\epsilon_{te}(t)$ during these four phases can be characterized by two timescales: $t_m$ -- the time when the test error reaches its minimum and $t_f$ -- the time when the training loss function reaches its minimum, and the two corresponding test errors: $\epsilon_m$ and $\epsilon_f$. All four parameters depend on $\rho$.  As shown in Fig.~\ref{rho}(C), $t_m$ is almost independent of $\rho$, which means that learning the correctly labeled data is independent of data size as long as the data size is large enough. However, $t_f$ increases with $\rho$, which means that the network needs more time to memorize the incorrectly labeled data as the number of mislabeled samples increases. As shown in Fig.~\ref{rho}(D), the final test error $\epsilon_{f}$ increases with $\rho$ almost linearly, which is caused by the increased fraction of mislabeled data. The minimum error $\epsilon_m$ remains roughly the same when $\rho$ is small, but increases sharply after a threshold and approaches $\epsilon_f$ when $\rho >0.85$. This also makes sense because when $\rho$ is large, learning is dominated by mislabeled data and the correctly labeled data no longer drives the learning dynamics.

\FloatBarrier

Here, we try to understand the different phases and the transitions between them by using order parameters that are modified for the case with labeling noise. In particular, each minibatch $\mu$ now consists of two smaller minibatches $\mu_c$ and $\mu_w$ for the correctly and incorrectly labeled data  ($\mu =\mu_c +\mu_w$) with the average size $B_c=(1-\rho)B$ and $B_w =\rho B$  respectively. The minbatch loss function can be decomposed into two minibatch loss functions $L^{\mu_c}$ and $L^{\mu_w}$ defined for $\mu_c$ and $\mu_w$ separately: $L^\mu =L^{\mu_c}+L^{\mu_w}$. At a given point in weight space, the ensemble averaged gradient and activity for the correctly and incorrectly labeled data can be defined separately:
\begin{eqnarray}
    g_c &\equiv & \langle \frac{\partial L^{\mu_c}}{\partial w}\rangle _{\mu_c}=\frac{\partial L_c}{\partial w}\;\;, \;\; A_c\equiv \langle \|\frac{\partial L^{\mu_c}}{\partial w}\|^2\rangle _{\mu_c},\\
    g_w &\equiv & \langle \frac{\partial L^{\mu_w}}{\partial w}\rangle _{\mu_w}=\frac{\partial L_w}{\partial w}\;\;, \;\; A_w\equiv \langle \|\frac{\partial L^{\mu_w}}{\partial w}\|^2\rangle _{\mu_w}.
\end{eqnarray}
The alignment of the two gradients $g_c$ and $g_w$ can be characterized by the cosine of their relative angle:
\begin{equation}
    R_{cw}\equiv \frac{g_c\cdot g_w}{\|g_c\| \| g_w\|},
\end{equation}
from which we obtain the ensemble averaged gradient and activity for the whole dataset:
\begin{eqnarray}
    g&\equiv& \langle \frac{\partial L^{\mu}}{\partial w}\rangle _{\mu} =(1-\rho)g_c +\rho g_w,\\ A &\equiv& \langle \|\frac{\partial L^{\mu}}{\partial w}\|^2\rangle _{\mu} =(1-\rho)^2 A_c +\rho^2 A_w +2\rho (1-\rho) \|g_c\| \| g_w\| C_{cw}.
\end{eqnarray}
From these basic ordered parameters defined above, we can define the directed activity $A_{0,c}\equiv (1-\rho)^2\|g_c\|^2$, $A_{0,w}\equiv \rho^2 \|g_w\|^2$, and $A_0 \equiv \|g\|^2 = A_{0,c} + A_{0,w} +2 [A_{0,w}A_{0,c}]^{\frac{1}{2}} C_{cw}$; and the alignments between $g$ and $g_c$, and between $g$ and $g_w$ are: $R_{aw}\equiv \frac{g \cdot g_w}{\|g\| \| g_w\|}$, $R_{ac}\equiv \frac{g \cdot g_c}{\|g\| \| g_c\|}$. We can also define alignment order parameters among members within the different gradient ensembles ($\{\mu_c\}$, $\{\mu_w\}$, and $\{\mu\}$).

\begin{figure}[ht]
\centering
\includegraphics[width = 0.75\linewidth]{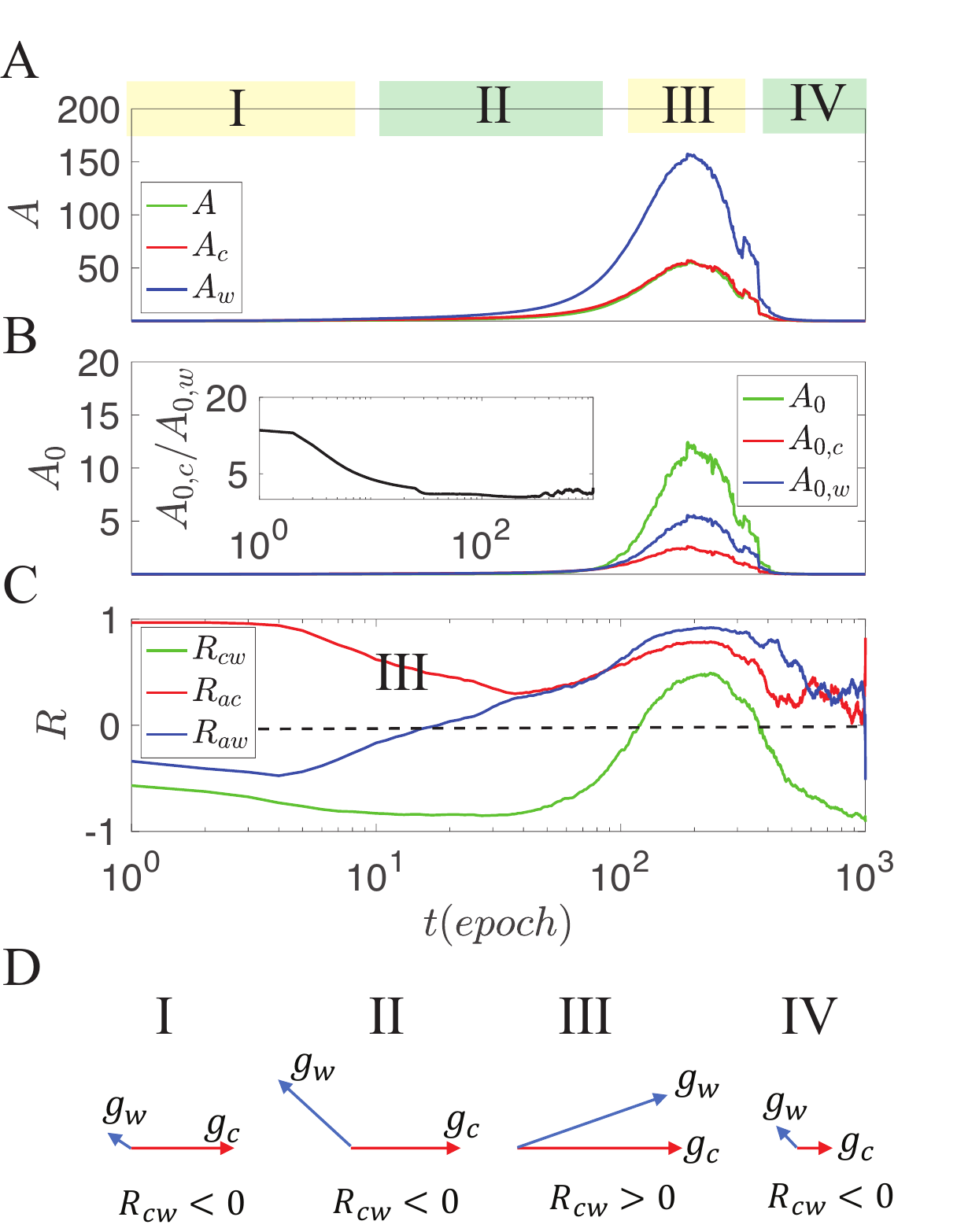} 

\caption{Dynamics of the order parameters during phases of learning with mislabeled data. (A) The total activities ($A$, $A_w$, $A_c$). (B) Directed activities ($A_0$, $A_{0,w}$, $A_{0,c}$), the inset shows the ratio $A_{0,c}/A_{0,w}$. (C) Alignment parameters ($R_{cw}$, $R_{ac}, R_{aw}$). The dotted line shows $R=0$. (D) Illustration of the four different phases in terms of the relative strength and direction of the two mean gradients ($g_c$ and $g_w$). }
\label{op_n}
\end{figure}

We studied three groups of order parameters: the total activities ($A$, $A_c$, $A_w$); the directed activities ($A_0$, $A_{0,c}$, $A_{0,w}$) and their alignments ($R_{cw}$, $R_{aw}$, $R_{ac}$) to understand the learning dynamics in the presence of labeling noise. As shown in Fig.~\ref{op_n}(A)\&(B), all learning activity order parameters ($A$'s and $A_0$'s) show a consistent trend of increasing during phase I, II, and III before deceasing during phase IV. This is in contrast to the behavior of learning activity $A$ in the absence of labeling noise, which shows a relatively flat or a slight decreasing trend during the fast learning phase (see Fig.~\ref{pt}). This continuously elevated learning activity in phases I-III suggests an increasing frustration between the two separate learning tasks (for learning the correctly and the incorrectly labeled datasets) before a consistent solution can be found in phase IV. 

The difference among learning phases I, II, and III can be understood by studying the relation between the two mean gradients $g_w$ and $g_c$ characterized by the alignment order parameter $R_{cw}$ (see Fig.~\ref{op_n}(C)) and the relative strength of the two directed activities $A_{0,c}$ and $A_{0,w}$. 
\begin{itemize}

\item Phase I: $A_{0,c}\gg A_{0,w}$, $R_{cw}<0$. In phase I, the directed activity from the correctly labeled data is much larger than that from the incorrectly labeled data (see inset in Fig.~\ref{op_n}(B)). This is due to the fact that samples from the correctly labeled dataset are consistent with each other in terms of their labels, which leads to a much larger mean gradient towards learning a solution for the correctly labeled data. In phase I, $g_c$ and $g_w$ are not aligned ($R_{cw}<0$). Due to the fact $A_{0,c}\gg A_{0,w} $, we have $R_{aw}<0$, which means that there is an increase of $L_w$ during phase I as observed in Fig.~\ref{rho}(A). 

\item Phase II: $A_{0,w}\approx A_{0,c}$, $R_{cw}< 0$. As the system approaches a solution for the correctly labeled data during late stage of phase I, the directed learning activity from the mislabeled data ($A_{0,w}$) increases sharply and $A_{0,w}$ become comparable with $A_{0,c}$ in phase II (see inset of the middle panel in Fig.~\ref{op_n}). In addition, the two mean gradients ($g_c$ and $g_w$) are opposite to each other with $R_{cw}\approx -1$. As a result of the balanced gradients between the two datasets, the overall directed activity is small $A_0 \ll A_{0,c(w)}$ and the loss functions ($L_c$, $L_w$, and $L$) remains relatively flat during phase II (see Fig.~\ref{rho}(A)). 

\item Phase III: $A_{0,w}\approx A_{0,c}$, $R_{cw}>0$. The system enters into phase III when it finally finds a direction to decrease both loss functions ($L_w$ and $L_c$) as evidenced by the alignment of $g_c$ and $g_w$, which only happens during phase III. This alignment ($R_{cw}>0$) means that the system can finally learn a solution for all the training data.

\item Phase IV: $A_{0,w}\approx A_{0,c}$, $R_{cw}<0$. Once the system finds a solution for all data, learning slows down to explore other solutions nearby. Phase IV is similar to the exploration phase without mislabeled data where learning activity is much reduced than those in phases I-III. 

\end{itemize}

Key differences of the four phases in terms of the strength and relative direction of the two mean gradients ($g_c$ and $g_w$) are illustrated in Fig.~\ref{op_n}(D).

\FloatBarrier

\begin{figure}[htbp]
\centering
\includegraphics[width=0.6\linewidth]{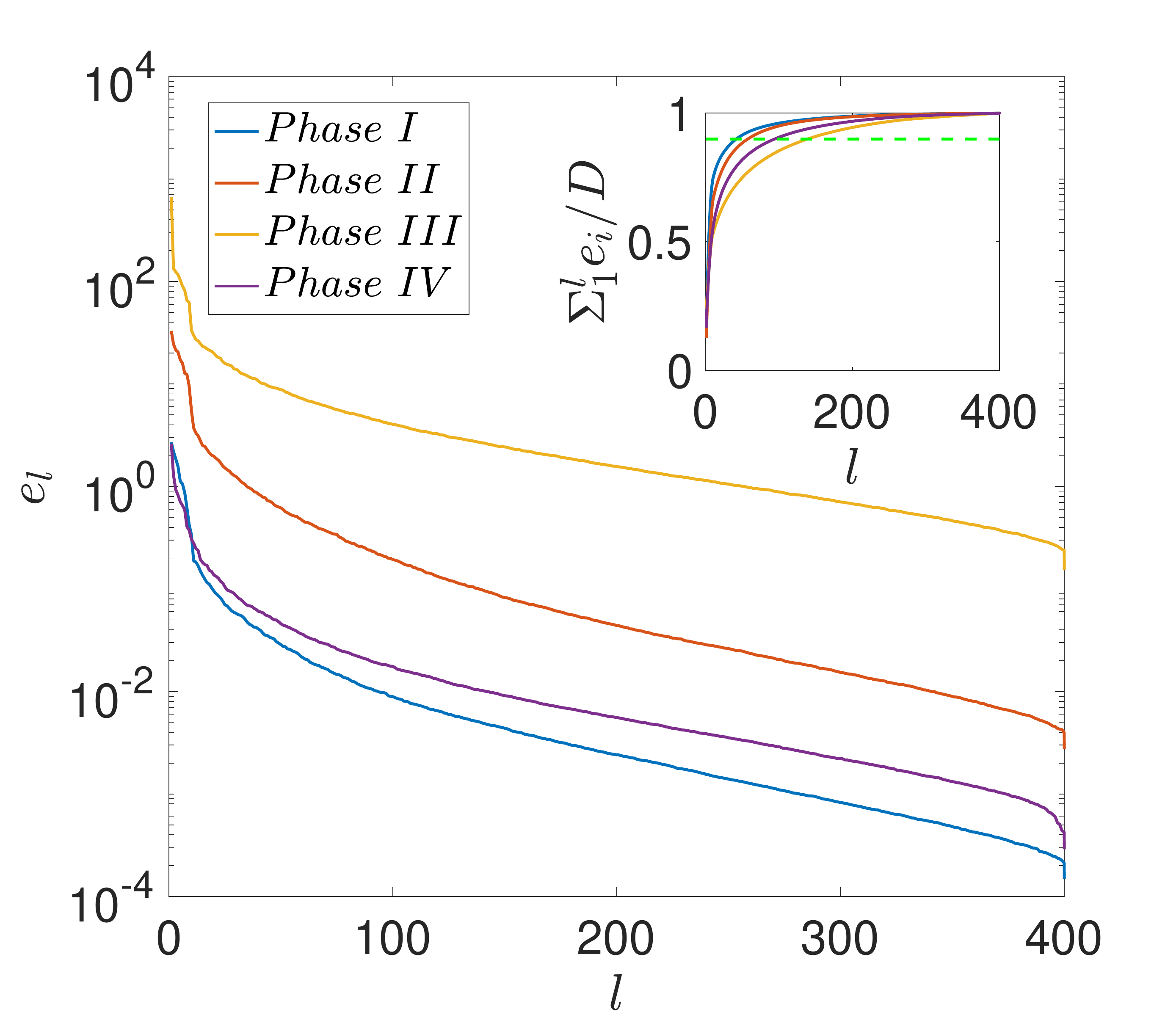}
\caption{The noise spectra, i.e., rank ordered eigenvalues $\{e_l,\; l=1,2..,N_p\}$ in different phases of learning with labeling noise (same setting as in Fig.~\ref{op_n}). The inset shows the normalized accumulated variance $D^{-1}\sum_{i=1}^{l} e_i$. The spectra are similar except for their total variance $D$. In different phases, the effective dimension $D_s$ varies in a range $(50-150)$, which is much smaller than the number of parameters ($N_p=2500$). } 
\label{noise_spec_n} 
\end{figure}

We have also analyzed the noise spectra in different learning phases in the presence of labeling noise. As shown in Fig.~\ref{noise_spec_n}, the normalized spectra remain roughly the same in different learning phases and the effective dimensions are $D_{I,II,III,IV}\approx 43,58,140, 95$, which are much smaller than the number of parameters. We note that both the noise spectra and the effective noise dimensions are similar to those without labeling noise (Fig.~\ref{noise_spec}).

\section{Identifying and cleaning the mislabeled samples in phase II}

\begin{figure}[ht]
\centering
\includegraphics[width = 0.85\linewidth]{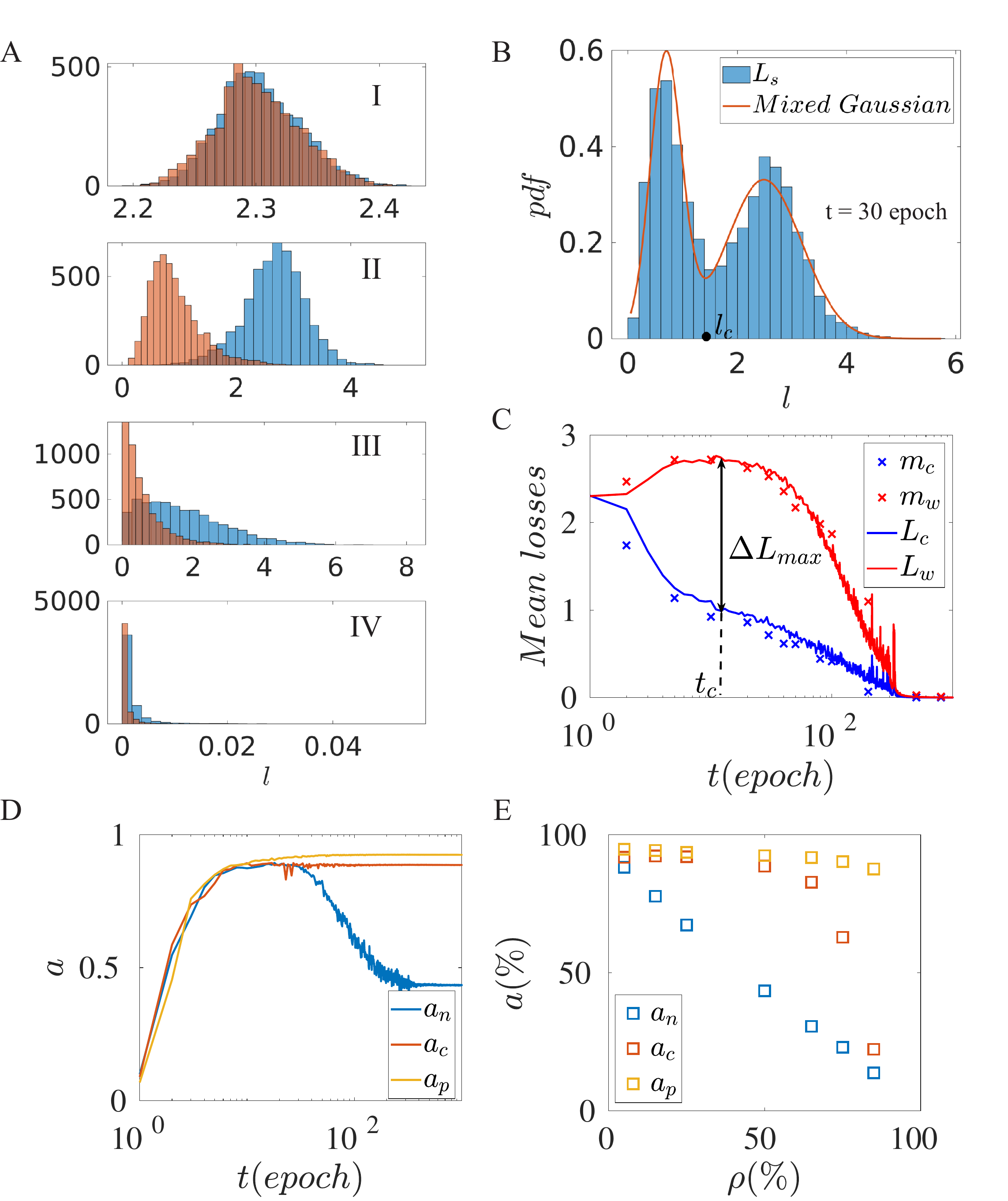} 

\caption{The individual loss distribution and the cleaning method. (A) The loss distributions of correctly labeled samples (red) and mislabeled samples (blue) in different learning phases. (B) The bimodal distribution in phase II can be fitted by a Gaussian mixture model (red line), which is used to determine a threshold $l_c$ for cleaning. (C) The mean losses (symbols) predicted from the Gaussian mixture model agree with their true values from experiments (lines). A cleaning time $t_c$ can be determined when $\Delta L(\equiv m_w-m_c)$ reaches its maximum. (D) The test accuracy without cleaning ($a_n$), with cleaning ($a_c$), and with only the correctly labeled training data ($a_p$) versus training time. The labeling noise level $\rho=50\%$ for (A)-(D). (E) $a_n$, $a_c$, and $a_p$ versus $\rho$. The slight decrease in $a_p$ as $\rho$ increases is due to the decreasing size of the correctly labeled dataset. MNIST dataset and network used here are the same as those in Fig.~\ref{rho}. }
\label{pdf}
\end{figure}

Our study so far has used various ensemble averaged properties to demonstrate the different phases of learning dynamics. We now investigate the distribution of losses for individual samples and how the individual loss distribution evolves 
with time. In Fig.~\ref{pdf}(A), we show the probability distribution functions (pdf's) - $P_c(l,t)$ and $P_w(l,t)$ - for the individual losses of the correctly labeled and incorrectly labeled samples at different times during training. Starting with an identical distribution at time $0$, the two distributions quickly separate during phase I as $P_c(l,t)$ moves to smaller losses while $P_w(l,t)$ moves slightly to higher losses. The separation between the two distributions increases during phase I and reaches its maximum during phase II. After the system enters phase III, the gap between the two distributions closes quickly as the system learns the mislabeled data and $P_w(l,t)$ catches up with $P_c(l,t)$ at small losses. In phase IV, these two distributions becomes indistinguishable again as they both become highly concentrated at near zero losses. 

As a result of the different dynamics of the two distribution, the overall individual loss distribution $P(l)=(1-\rho)P_c(l)+\rho P_w(l)$ exhibits a bimodal behavior, which is most pronounced during phase II. In fact, we can fit the overall distribution by a Gaussian mixture model: $l\sim (1-r) \mathcal{N}(m_c,s_c^2) + r \mathcal{N}(m_w,s_w^2)$ with fitting parameters: fraction $r$, means $m_{c,w}$, and variances $s^2_{c,w}$. As shown in Fig.~\ref{pdf}(B), the Guassian mixture model fits $P(l)$ well, and furthermore, the fitted means $m_c$ and $m_w$ agree with the mean losses ($L_c$, and $L_w$) obtained from the experiments. % over all phases of learning. %  Without knowing the identity of each individual sample  Here, we describe the distribution of the loss for individual training sample and the bimodal distribution during phase II and how we use the insight to clean the training data and obtain better test accuracy. The improvement depends on $\rho$ etc. We should show the distribution at different time and how they become bimodal during phase II, and how cleaning help improve test accuracy for MNIST and CIFAR.

The separation of individual loss distribution functions has recently been used to devise sophisticated methods to improve generalization such as those reported in ~\cite{arazo2019unsupervised,Li2020GradientDW}. Here, we demonstrate the basic idea by presenting a simple method to identify and clean the mislabeled samples based on the understanding of different learning phases. In particular, according to our analysis, such a cleaning process can be best done during phase II. For simplicity, we set the time $t_c$ for cleaning when the difference $\Delta L(\equiv m_w-m_c)$ reaches its maximum. At $t=t_c$, we can set a threshold $l_c$, which best separates the two distributions. For example, we can set $l_c$ as the loss when the two pdf's are equal or simply as the average of $m_c$ and $m_w$ (we do not observe significant differences between the two choices). We can then get rid of all the data which has a loss larger than $l_c$ and continue training with the cleaned dataset. Alternatively, we can stop the training altogether at $t=t_c$, i.e., early stopping. We do not observe significant differences between these two choices in our experiments. In Fig.~\ref{pdf}(D), the test accuracy $a_n$ without cleaning, $a_c$ with cleaning, and $a_p$ with only the correctly labeled data are shown for MNIST data with $\rho=50\%$ labeling noise. Performance of the cleaning algorithm can be measured by $Q = \frac{a_c-a_n}{a_p-a_n}$, which depends on the noise level $\rho$. As shown in Fig.~\ref{pdf}(E), the cleaning method can achieve significant improvement in generalization ($Q>50\%$) for noise level up to $\rho=80\%$ noise level.

\section{Summary}

Deep learning neural networks have demonstrated tremendous capability in learning and problem solving in diverse domains. Yet, the mechanism underlying this seemingly magical learning ability is not well understood. For example, modern DNNs often contain more parameters than training samples, which allow it to interpolate (memorize) all the training samples, even if their labels are replaced by pure noise~\cite{zhang2016understanding, arpit2017closer}. Remarkably, despite their huge capacity, DNNs can achieve small generalization error on real data (this phenomenon has been formalized in the so called  ``double descent" curve~\cite{Belkin2019DoubleDescent, brutzkus2017sgd,li2018learning, mei2019generalization, geiger2020scaling, gerace2020generalisation}). The learning system/model seems to be able to self-tuned its complexity in accordance with the data to find the simplest possible solution in the highly over-parameterized weight space. However, how does the system adjusts its complexity dynamically, and how SGD seeks out simple and more generalizable solutions for realistic learning tasks remain not well understood.     %a careful study of the SGD dynamics and the loss function landscape in this paper reveals a robust inverse relation between fluctuations in SGD and flatness of the loss landscape, which is critical for deciphering the learning strategy in deep learning and for designing more efficient algorithms. 

In this paper, we demonstrate that the approach based on statistical physics and stochastic dynamical systems provides a useful theoretical framework (alternative to the traditional theorem proving approach) for studying SGD-based machine learning by applying it to identify and characterize the different phases in SGD-based learning with and without labeling noise.  In an earlier work~\cite{feng2020neural}, we have used this approach to study the relation between SGD dynamics and the loss function landscape and discovered an inverse relation between weight variance and the loss landscape flatness that is the opposite to fluctuation-dissipation relation (the Einstein relation) in equilibrium systems. We believe this framework may pave the way for a deeper understanding of deep learning by bringing powerful ideas (e.g., phase transitions in critical phenomena) and tools (e.g., renormalization group theory and replica method) from statistical physics to bear on understanding ANN. It would be interesting to use this general framework to address other fundamental questions in machine learning such as generalization~\cite{Neyshabur2017ExploringGI,advani2017highdimensional, jiang2019fantastic} in particular the mechanism for the double descent behavior in learning as described above; the relation between task complexity and network architecture; information flow in DNN~\cite{shwartz2017opening,Tishby_2015}; as well as building a solid theoretical foundation for important applications such as transfer learning~\cite{yosinski2014transferable}, curriculum learning~\cite{bengio2009curriculum}, and continuous learning~\cite{Ring94,GEM,riemer2018learning}.

%\bibliography{ML}

\end{document}